\documentclass[10pt,twocolumn,letterpaper]{article}

\usepackage{cvpr}              
\definecolor{cvprblue}{rgb}{0.21,0.49,0.74}
\usepackage[pagebackref,breaklinks,colorlinks,allcolors=cvprblue]{hyperref}
\usepackage{colortbl}
\usepackage{amssymb}
\usepackage{pifont}
\newcommand{\cmark}{\ding{51}}%
\newcommand{\xmark}{\ding{55}}%

\def\paperID{7429} 
\def\confName{CVPR}
\def\confYear{2026}

\title{Deformable Gaussian Occupancy: Decoupling Rigid and Nonrigid Motion with Factorized Distillation}

\author{Yang Gao \quad Wuyang Li \quad Po-Chien Luan \quad Alexandre Alahi\\
\'{E}cole Polytechnique F\'ed\'erale de Lausanne (EPFL), Switzerland\\
{\tt\small \{firstname.lastname\}@epfl.ch}
}

\begin{document}

\maketitle
\let\thefootnote\relax\footnotetext{\leftline{Correspondence to: Wuyang Li and Alexandre Alahi}}

\begin{abstract}
Understanding dynamic 3D environments is essential for safe autonomous driving, particularly when reasoning about human-centric, nonrigid agents. However, existing weakly supervised occupancy prediction frameworks predominantly assume rigid-body motion and rely on simple frame-to-frame offsets, limiting their ability to capture fine-grained deformations and maintain temporal coherence. To address this issue, we propose \textbf{DeGO}, a deformable Gaussian occupancy framework that unifies decoupled Gaussian deformation with factorized 4D foundation-model distillation. DeGO disentangles rigid and nonrigid motion, enabling each Gaussian primitive to evolve through both deformation and offset-based updates. In parallel, a factorized 4D distillation strategy transfers cross-camera and cross-frame knowledge from the VGGT foundation model, producing foundation-aligned features that enhance temporal consistency. Experiments on the Occ3D-NuScenes benchmark demonstrate that our method achieves state-of-the-art performance under weak supervision, delivering 13.5\% gains on human-centric instances and 10.9\% overall improvements. These results highlight the effectiveness of deformation-aware and foundation-guided occupancy modeling for dynamic scene understanding. 
The code is publicly available: \href{https://github.com/vita-epfl/DeGO}{\color{magenta}https://github.com/vita-epfl/DeGO}.

\end{abstract}    
\section{Introduction}
\label{sec:intro}

Dynamic 3D occupancy prediction has become a key capability for camera-based autonomous systems, providing a unified view of scene geometry, semantics, and motion. Yet, existing methods struggle to scale because voxel labels are costly to obtain, and most camera-only models infer complex dynamics from limited supervision.
These challenges are amplified in dynamic environments, where rigid and nonrigid motions introduce temporal inconsistencies that prior voxel or Gaussian approaches fail to model.

To address this bottleneck, recent work has explored weakly supervised learning for occupancy prediction, reducing reliance on costly manual voxel labels. Some methods~\cite{huang2024selfocc,boeder2025gaussianflowocc} exploit inherent temporal consistency to provide cross‑frame supervision and enforce 3D consistency. With the emergence of efficient 3D reconstruction and rendering techniques such as Gaussian Splatting (GS)~\cite{kerbl20233d} and  Neural Radiance Field (NeRF)~\cite{mildenhall2021nerf}, recent approaches~\cite{jiang2025gausstr,gan2025gaussianocc,boeder2025gaussianflowocc} learn feed‑forward GS models to represent scenes and optimize rendered 2D features~\cite{jiang2025gausstr} or semantic maps~\cite{boeder2025gaussianflowocc}. Because of their efficiency and strong performance, these methods have recently become the dominant solution.

\begin{figure}
    \centering
    \includegraphics[width=\linewidth]{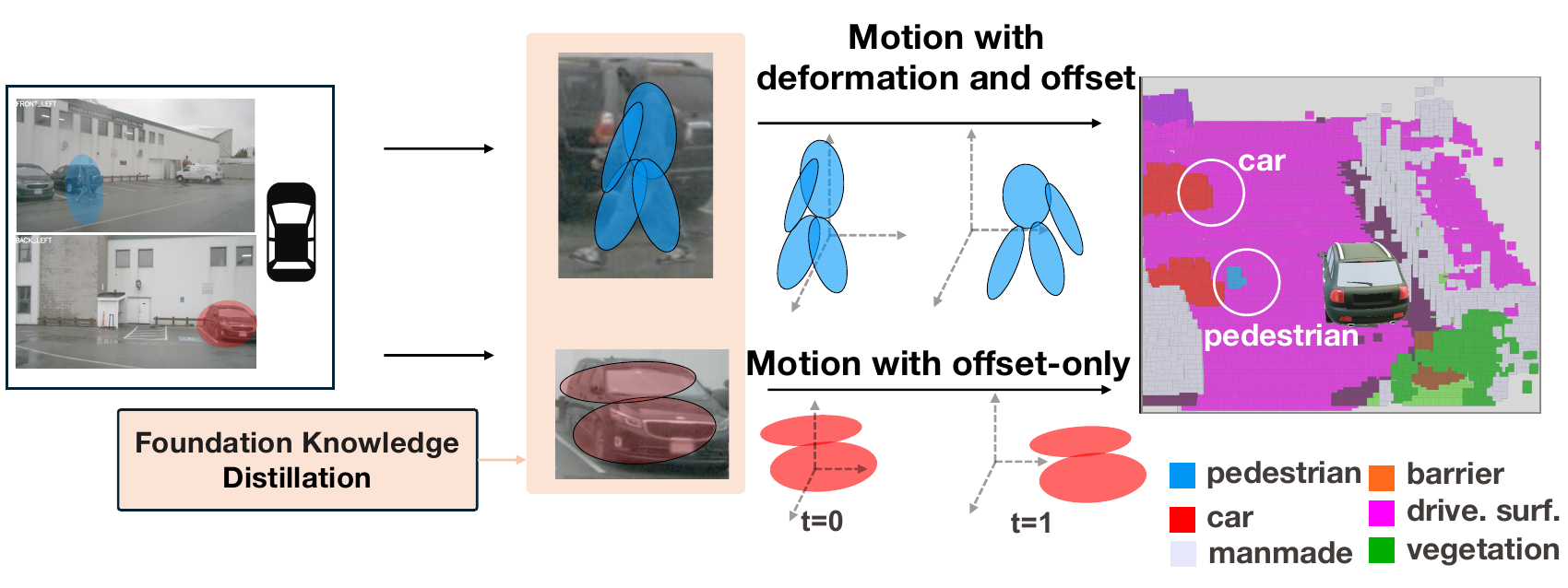}
    \caption{\textbf{Overview of our deformable Gaussian occupancy framework.} We enable Gaussians to adaptively model rigid and nonrigid motion. Deformable Gaussians evolve through both nonrigid deformation and offsets, while rigid Gaussians use only offset updates. Foundation-model distillation provides cross-camera and cross-frame guidance, yielding more accurate occupancy prediction via temporal consistency.}
    \label{fig:intro}
\end{figure}

Despite recent advances, current methods remain \textbf{fundamentally limited in their ability to model human-centric dynamic scenes}. Because they distribute Gaussian primitives uniformly across the 3D volume, most capacity is consumed by large, static background regions (e.g., roads, walls), leaving insufficient modeling resolution for small yet safety-critical human instances. At the same time, the underlying motion model is constrained to rigid translations or simple frame-to-frame offsets~\cite{boeder2025gaussianflowocc}, preventing these methods from capturing the fine-grained nonrigid deformations that characterize human motion~\cite{luan2025unified,hosseininejad2025motionmap,tokoro2026mmcm}. This combination of capacity imbalance and rigid-motion assumptions leads to under-representation of human geometry, degraded human-related mIoU, and unreliable temporal consistency. 

Recent progress in dynamic scene rendering has demonstrated the importance of deformable GS~\cite{wu20244d,yang2023real} for non-rigid motions. These works highlight a key insight: representing nonrigid dynamics requires an explicit mechanism to deform spatial primitives over time. \emph{While promising solutions, directly applying them to occupancy prediction is non-trivial due to two key issues}. \textbf{First}, driving scenes contain a heterogeneous mix of rigid structures and highly nonrigid human agents, making a single unified deformation field inappropriate. Without explicitly separating rigid and nonrigid motion, deformable GS entangle incompatible dynamics and produce unstable geometry updates. \textbf{Second}, deformable GS is limited to weakly supervised optimization. These challenges motivate our decoupled deformation formulation and our use of foundation-model distillation to provide stable 4D guidance.

To address these challenges, we propose the \textbf{Deformable Gaussian Occupancy (DeGO)} framework with 4D foundational guidance (see \Cref{fig:intro}) for weakly supervised occupancy prediction. Rather than treating classes uniformly~\cite{gan2025gaussianocc,jiang2025gausstr,boeder2025gaussianflowocc,huang2024selfocc}, DeGO enables each Gaussian to \emph{adaptively model its motion behavior}. Specifically, we introduce \textbf{Decoupled Gaussian Deformation (DGD)}, a Gaussian parameterization that predicts a soft rigidity mask to control whether a Gaussian undergoes rigid motion, nonrigid deformation, or a combination of both. This flexible formulation allows the model to capture hybrid dynamics while preserving geometric stability in rigid regions.
To further enhance temporal coherence and multi-view consistency, we incorporate \textbf{Factorized Feature Distillation (FFD)}, which transfers 4D spatiotemporal knowledge from foundation models such as VGGT into the Gaussian field. Together, these components yield a deformation-aware and foundation-aligned 4D representation. 
Extensive experiments show that DeGO can not only surpass state-of-the-art works with 10.9\% mIoU, but also shows a 13.5\% gain on the human-centric metric. Our contributions can be summarized as follows:

\begin{enumerate} 

    \item We present a unified deformable Gaussian occupancy framework that enables each Gaussian to adaptively model rigid and nonrigid motion. Guided by 4D foundation models, our approach improves both human-centric and overall scene understanding.
    
    \item We propose the human-centric decoupled Gaussian deformation, which allocates deformation capacity selectively, enabling accurate and temporally consistent modeling of both nonrigid and rigid motion.

    \item We introduce a factorized 4D feature distillation that transfers cross-camera and cross-frame knowledge from the foundation model to the Gaussian field, producing foundation-aligned dynamic scene representations.

\end{enumerate}

\section{Related Works}
\label{sec:related_works}

\subsection{Fully Supervised 3D Occupancy Prediction}
Evolving from semantic scene completion task~\cite{song2017semantic,cao2024pasco}, 3D occupancy prediction aims to recover both the scene geometry and the semantics within a voxel representation. Fully supervised methods have progressed rapidly with datasets providing dense voxel annotations. Early methods~\cite{cao2022monoscene, zhang2023occformer, huang2023tri, li2023voxformer,tong2023scene,huang2021bevdet,li2024bevformer,jiang2024symphonize,mei2024camera} project 2D features into voxel grids using depth or lifted BEV feature, but they depend on dense 3D labels and suffer from heavy computational costs due to the cubic scaling of voxel resolution.
Although effective, dense grids incur significant computational overhead due to the large volume of empty space. VoxDet~\cite{li2025voxdet} reinterprets occupancy prediction as dense detection, deriving instance-aware supervision.

To improve efficiency, recent approaches~\cite{liu2024fully,lu2024octreeocc,shi2024occupancy,tang2024sparseocc,wang2024opus,yu2023flashocc,huang2024gaussianformer} adopt sparse and adaptive scene representations while retaining full 3D supervision. GaussianFormer~\cite{huang2024gaussianformer} represents scenes using sparse Gaussians and learns geometry and semantics through Gaussian-to-voxel splatting. GaussianFormer-2~\cite{huang2024probabilistic} introduces a probabilistic Gaussian superposition formulation, improving efficiency by explicitly modeling only non-empty space. VoxelSplat~\cite{zhu2025voxelsplat} introduced the rigid motion to the Gaussians, enabling better temporal consistency. While compact, these methods remain fully supervised and rely on expensive voxel labels.

Overall, fully supervised methods, which encompass grid-based, Gaussian-based, and detection-based approaches, demonstrate strong performance by leveraging accurate voxel-level labels. However, this reliance on dense 3D annotations makes them costly to train at scale.

\subsection{Weakly Supervised 3D Occupancy Prediction}

Weakly supervised approaches aim to learn 3D occupancy without relying on dense voxel annotations, instead leveraging 2D cues such as depth, semantics, or multi-view consistency. Early weakly supervised methods~\cite{huang2024selfocc, zhang2025occnerf, wang2024distillnerf} reconstruct 3D geometry through differentiable rendering or by distilling 3D priors from foundation features. While label-efficient, these methods inherit the computational burden of dense voxel grids and struggle to maintain temporal consistency in dynamic scenes.

Recent Gaussian-based frameworks~\cite{gan2025gaussianocc, jiang2025gausstr, boeder2025gaussianflowocc, chambon2025gaussrender} address these limitations by replacing voxel grids with sparse Gaussian primitives that can be rendered differentiably. GaussianOcc~\cite{gan2025gaussianocc} accelerates optimization with compact Gaussians; GaussRender~\cite{chambon2025gaussrender} links image-space supervision to 3D structure through Gaussian rendering; GaussTR~\cite{jiang2025gausstr} aligns Gaussian features with foundation models~\cite{oquab2023dinov2} for open-vocabulary reasoning; VEON~\cite{zheng2024veon} and LangOcc~\cite{boeder2024langocc} use vision-language features to augment the voxel renderings; and GaussianFlowOcc~\cite{boeder2025gaussianflowocc} introduces per-Gaussian temporal offsets for dynamic scenes and link the rendered Gaussians with pseudo semantic labels from the GroundedSAM~\cite{ren2024grounded} and depth labels from the Metric3D~\cite{yin2023metric3d}.

Despite their efficiency, current Gaussian-based weakly supervised models typically assume rigid-body motion and use simple frame-to-frame offsets, making them insufficient for modeling human-centric nonrigid motions. Our work extends this direction by introducing a decoupled 4D Gaussian deformation that separates rigid and nonrigid motion and complements it with factorized spatiotemporal distillation, achieving consistent and deformation-aware weakly supervised occupancy prediction.

\section{Preliminary and Motivation}
\label{sec:preliminary}

\noindent\textbf{Problem Formulation.} 
Given a sequence of multi-view images $\mathcal{I}_{-T:0} = \{ \mathcal{I}_t^{(v)} \mid v \in \mathcal{V},\; t=-T,\dots,0 \},$
captured across $\mathcal{V}$ synchronized cameras, our goal is to reconstruct a temporally consistent 3D scene representation that jointly encodes geometry and semantics. Formally, we learn a mapping from the 3D world system to the voxel system
\begin{equation}
    f_\theta: (\mathbf{x}, t) \mapsto (p_{\text{occ}}, p_{\text{sem}}), 
    \quad \mathbf{x} \in \mathbb{R}^3,\; t \in [-T, 0],
\end{equation}
where $\mathbf{x}$ is 3D spatial coordinate in world system, $p_{\text{occ}} \in [0,1]$ denotes the occupancy probability and 
$p_{\text{sem}} \in \Delta^{C}$ is the categorical distribution over $C$ semantic classes 
(e.g., humans, vehicles, vegetation).

To model this mapping efficiently, we discretize the continuous space into a voxel grid 
$\mathcal{V}_g \subset \mathbb{R}^3$ with voxel size $\delta$, 
and learn a latent \emph{feature volume}: $\mathbf{F}_t : \mathcal{V}_g \rightarrow \mathbb{R}^{C_f}$, 
where each voxel feature $\mathbf{f}_{x,t} \in \mathbb{R}^{C_f}$ encodes local geometry, semantics, 
and motion context at position $\mathbf{x}$ and time $t$.
The occupancy and semantic logits are then obtained by lightweight prediction heads:
\begin{align}
    p_{\text{occ}}(\mathbf{x}, t) &= \sigma(\mathbf{w}_{\text{occ}} \mathbf{f}_{x,t}), \\
    p_{\text{sem}}(\mathbf{x}, t) &= \mathrm{softmax}(\mathbf{w}_{\text{sem}} \mathbf{f}_{x,t}),
\end{align}
where $\mathbf{w}_{\text{occ}}$ and $\mathbf{w}_{\text{sem}}$ are learnable weights for their prediction heads, and $\sigma(\cdot)$ denotes the sigmoid function.

\noindent\textbf{Scene as Rigid 3D Gaussian.}\label{sec:pre_naive} 
To efficiently model continuous 3D geometry without relying on dense voxel volumes, 
GaussianFlow~\cite{boeder2025gaussianflowocc} adopts the GS representation with vanilla temporal module, which parameterizes the scene 
as a set of rigid Gaussian primitives and predict the offsets of each Gaussian. Mathematically, each scene at the current time step $t=0$ is represented by a set of $N$ learnable Gaussians
\[ \mathcal{G}_t = \{ G_i(t) = (\boldsymbol{\mu}_i(t),\, \mathbf{r}_i(t),\, \mathbf{s}_i(t),\, \mathbf{\alpha}_i(t) \}_{i=1}^{N}, \]
with $\boldsymbol{\mu}_i$ denoting the 3D position, $\mathbf{r}_i$ the rotation quaternion, $\mathbf{s}_i$ the scale, 
and $\mathbf{\alpha}_i$ the opacity of the $i$-th Gaussian.
When considering neighboring frames, each Gaussian $G_i(t)$ encodes 3D positions at different time steps $\boldsymbol{\mu}_i(t)$, but the orientation $\mathbf{r}_i(0)$, scaling $\mathbf{s}_i(0)$, and opacity $\mathbf{\alpha}_i(0)$ are determined by the reference frame (t=0). To connect the continuous Gaussian field with the discrete voxel representation, GaussianFlow~\cite{boeder2025gaussianflowocc} projects $\mathcal{G}_t$ onto the voxel grid $\mathcal{V}_g$ 
via a differentiable splatting operation, 
yielding voxel-aligned features $\mathbf{F}_t$ for occupancy and semantic prediction.

\noindent\textbf{Scene as Decoupled 4D Gaussian.} 
While such rigid 3D Gaussians have proven effective for diverse scenes, the assumption of a fixed shape per instance can only model motion 
through frame-to-frame offsets. 
This limitation makes it difficult to capture nonrigid deformations, 
especially for human-centric agents whose geometry changes over time~\cite{saadatnejad2024socialtransmotion,gao2024multitransmotion,gao2025social,gao2025omnitraj}. 
These challenges motivate our introduction of a deformation-aware 4D Gaussian representation that explicitly learns temporally coherent shape evolution.

Building upon the rigid representation above, 
we models dynamic scenes using a \emph{decoupled Gaussian deformation} that distinguishes 
rigid and nonrigid motion patterns in Gaussian feature space. 
Different from the prior work representing scene as rigid Gaussians, we first define that each Gaussian $G_i$ is augmented with a learnable rigid-body mask $m_i\!\in\![0,1]$, which implicitly controls whether it undergoes only rigid offsets 
or both offset and nonrigid deformation (see Sec~\ref{subsubsec:deform}). Secondly, we update the motion differently for rigid classes and nonrigid classes. Specifically, rigid classes are updated through simple offsets $\boldsymbol{\mu}_i(t)$, 
while nonrigid classes undergo additional local deformation (i.e., $\mathbf{r}_i(t),\, \mathbf{s}_i(t),\, \mathbf{\alpha}_i(t)$) driven by learned feature dynamics. 
This adaptive design allows the model to automatically allocate deformation capacity 
to nonrigid regions while preserving stability for rigid structures, 
forming the foundation for the subsequent deformation and distillation modules.

\section{Method}
\label{sec:method}

\noindent\textbf{Overview.} 
Given a short sequence of multi-view images, our framework models how the underlying 3D Gaussian scene representation evolves over time. During training, we sample both past and future frames and apply weak supervision with pseudo depth and segmentation, encouraging the model to maintain coherent geometry and semantics under motion. The temporal modeling acts purely as a training-time augmentation and does not bring extra inference cost. The framework integrates two modules: \textbf{(a) Decoupled Gaussian Deformation} module, which receives the evolving Gaussian features and predicts rigid and nonrigid motion separately, and a \textbf{(b) Factorized Feature Distillation} module, which injects multi-view and temporal cues from a foundation model into the Gaussian features before deformation. Together, they enable more reliable object motion and consistent scene geometry.

\begin{figure*}
    \centering
    \includegraphics[width=\linewidth]{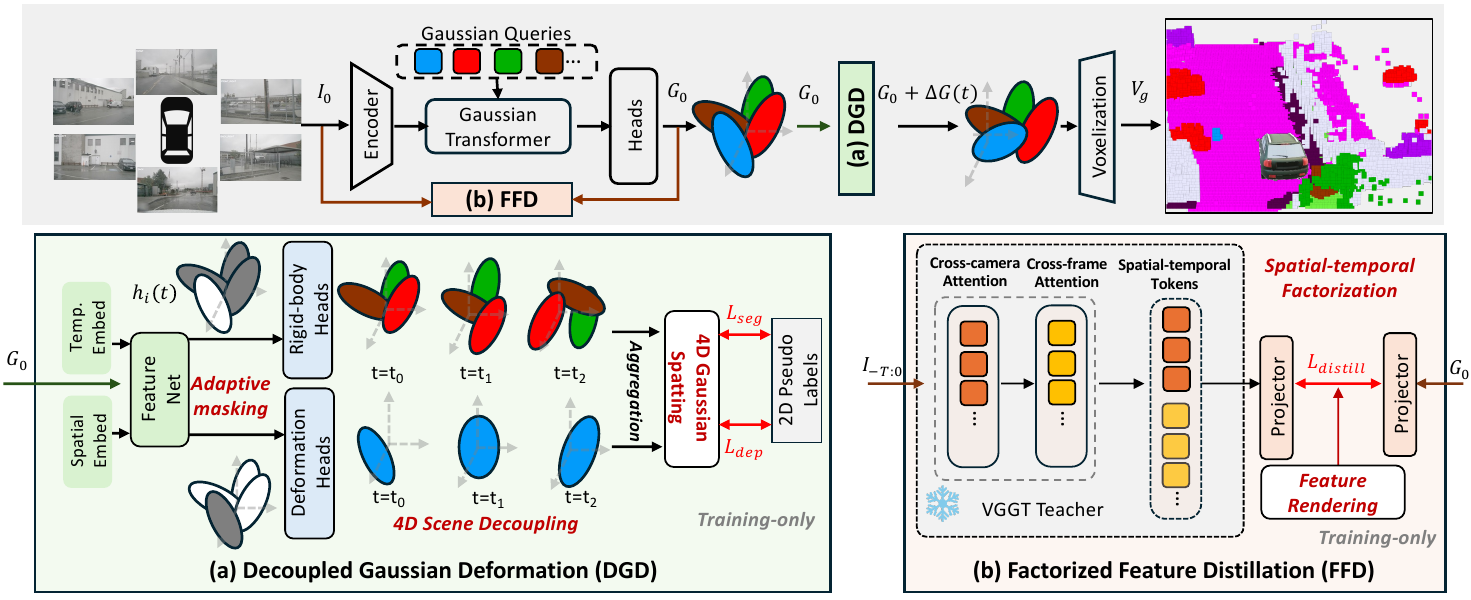}
    \caption{\textbf{Overview of the proposed DeGO framework.} 
    It unifies the Decoupled Gaussian Deformation (DGD) and Factorized Feature Distillation (FFD). The spatialtemporal features from VGGT teacher guides Gaussian rendering through feature alignment, producing foundation-aligned 4D features that drive decoupled motion prediction for nonrigid classes and rigid classes.}
    \label{fig:model}
\end{figure*}

\subsection{Decoupled Gaussian Deformation (DGD)}
\label{subsubsec:deform}

The goal of the deformation module is to model how each Gaussian evolves over time. 
Given the canonical Gaussians at the reference frame $t{=}0$, it predicts their 
position, rotation, scale, and opacity at $T$ selected target time steps $t\!\in\![-T,+T]$.
The deformation is guided by the foundation-aligned 4D features distilled from VGGT, 
which provide motion-aware cues to capture nonrigid dynamics across frames. 
This temporal deformation enables motion-aware training, while inference still uses a single frame.

\noindent\textbf{Input Representation.}
At time $t{=}0$, each Gaussian $G_i(0)$ has position $\boldsymbol{\mu}_i(0)$,
rotation quaternion $\mathbf{r}_i(0)$, scale $\mathbf{s}_i(0)$, 
opacity $\alpha_i(0)$, and latent feature $\mathbf{f}_i$ produced by the decoder.
For a scene with $N$ Gaussians each, the inputs are
\[
\boldsymbol{\mu}\!\in\!\mathbb{R}^{N\times3},\;
\mathbf{r}\!\in\!\mathbb{R}^{N\times4},\;
\mathbf{s}\!\in\!\mathbb{R}^{N\times3},\;
\alpha\!\in\!\mathbb{R}^{N\times1},\;
\mathbf{f}\!\in\!\mathbb{R}^{N\times C_g}.
\]
A sequence of time steps $\mathbf{t}\!\in\!\mathbb{R}^{T}$ 
defines the temporal frame offsets to be sampled during training.

\noindent\textbf{Temporal and Spatial Encoding.}
To capture smooth temporal variations, we apply positional encoding to both spatial coordinates and time:
\begin{align}
\gamma_p(\boldsymbol{\mu}) &= \big[\boldsymbol{\mu},\;\sin(2^k\boldsymbol{\mu}),\;\cos(2^k\boldsymbol{\mu})\big]_{k=0}^{L_p-1},\\
\gamma_t(t) &= \big[t,\;\sin(2^k t),\;\cos(2^k t)\big]_{k=0}^{L_t-1},
\end{align}
where $L_p$ and $L_t$ are the numbers of spatial and temporal frequency bands.
The encoded time $\gamma_t(t)$ is mapped by a small projector into a time embedding $\mathbf{e}_t\!\in\!\mathbb{R}^{C_t}$.
For each Gaussian and each timestamp, we concatenate its feature, encoded position, and encoded time:
\begin{equation}
\mathbf{h}_{i}(t)
=\mathrm{Feature Net}\big([\mathbf{f}_i,\ \gamma_p(\boldsymbol{\mu}_i(0)),\ \mathbf{e}_t]\big)
\in\mathbb{R}^{D_h},
\end{equation}
where $\mathrm{Feature Net}$ is a MLP with hidden dimension $D_h$.

\noindent\textbf{Deformation Heads and Rigid-body Mask.}
As illustrated in \Cref{fig:model}(a), each hidden Gaussian feature $\mathbf{h}_i(t)$ is fed into a set of lightweight MLP deformation heads that predict motion updates for every Gaussian. To decouple rigid and nonrigid behavior, we introduce a learnable \emph{adaptive rigid-body mask} $m_i\!\in\![0,1]$ indicating the degree of rigidity, 
where $m_i\!\approx\!1$ for nonrigid regions and $m_i\!\approx\!0$ for rigid ones. 
Two types of motion updates are then produced: a \emph{rigid offset} $\Delta G_i^{\text{rig}}(t)$ 
and a \emph{nonrigid deformation} $\Delta G_i^{\text{def}}(t)$ predicted only when $m_i$ is high. 
The combined update is expressed as
\begin{equation}
    \Delta G_i(t) = (1 - m_i)\,\Delta G_i^{\text{rig}}(t) + m_i\,\Delta G_i^{\text{def}}(t).
\end{equation}

A binary regularization loss is imposed on the rigid-body mask to encourage near-binary outputs, 
driving each Gaussian toward either rigid or nonrigid behavior. 
Consequently, rigid instances receive only positional offsets, 
whereas nonrigid ones undergo both offset and deformation. 
This adaptive control preserves geometric stability for rigid structures 
while enabling flexible modeling of nonrigid motion.

\noindent\textbf{Deformation Loss.}
We regularize the deformation to ensure temporal smoothness:
$$
\mathcal{L}_{\mathrm{reg}}
=\!\!\sum_{p\in\{\mu,r,s,\alpha\}}\!\!\lambda_p\,\|\Delta p_i(t)\|_2^2,
\quad
\mathcal{L}_{\mathrm{mask}}=[\mathbf{m}_i(1-\mathbf{m}_i)].
$$
The full deformation loss is
\begin{equation}
\mathcal{L}_{\mathrm{def}} = \lambda_{\mathrm{reg}}\mathcal{L}_{\mathrm{reg}} + \lambda_{\mathrm{mask}}\mathcal{L}_{\mathrm{mask}}.
\end{equation}

\subsection{Factorized Feature Distillation (FFD)}
\label{subsubsec:vggt}
Unlike previous approaches that distill per-frame or image-space features from 2D teachers (\eg, DINO or CLIP), we introduce a \emph{Factorized Feature Distillation (FFD)} to exploit VGGT’s spatialtemporal representations as 4D supervisory signals. As the teacher model, VGGT jointly encodes cross-camera and cross-frame context, providing richer structural cues. We then align rendered Gaussian features with 4D teacher features, enabling the model to refine each Gaussian’s predicted position, rotation, scale, and opacity over time.
By transferring the teacher’s global multi-view and temporal reasoning ability into the Gaussian field, FFD improves dynamic scene understanding while maintaining the inference efficiency of the Gaussian feedforward backbone.

\noindent\textbf{Decomposed 4D Foundational Feature.}
We first resize each image to multiples of the patch size $P$ and feed the multi-view sequence to VGGT's aggregator. 
Internally, tokens are formed per view and per time step by a patch embedding, with additional camera and register tokens. 
As \Cref{fig:model} (b) shows, we then process alternates \textbf{cross-camera (spatial)} attention and \textbf{cross-frame (temporal)} attention at each transformer block. For block $b$,
\begin{align}
\text{(spatial)}\quad 
\tilde{\mathbf{z}}^{(b)}_{t,v} &= \mathrm{Attn}_{\text{sp}}\!\Big(\mathbf{z}^{(b-1)}_{t,v}, \{\mathbf{z}^{(b-1)}_{t,u}\}_{u\in\mathcal{V}}\Big), \\
\text{(temporal)}\quad 
\hat{\mathbf{z}}^{(b)}_{t,v} &= \mathrm{Attn}_{\text{tmp}}\!\Big(\tilde{\mathbf{z}}^{(b)}_{t,v}, \{\tilde{\mathbf{z}}^{(b)}_{\tau,v}\}_{\tau\in[-T,0]}\Big).
\end{align}
We used the interleaved implementation: each block first fuses tokens across cameras, then across time. 
We refer to the teacher features by $\mathbf{T}^{(\ell)}_t(v)$ at selected block index $\ell\!\in\!\mathcal{L}$, where $\mathcal{L}$ is the total number of blocks of VGGT.

For a selected block $\ell\!\in\!\mathcal{L}$ and camera $v$ at the reference time $t\!=\!0$, we take both outputs of that block:
the spatial-output tokens $\tilde{\mathbf{z}}^{(\ell)}_{0,v}$ and the temporal-output tokens $\hat{\mathbf{z}}^{(\ell)}_{0,v}$.
After removing the camera and register tokens, we reshape each set back to feature maps $\mathbf{T}^{(\ell,\mathrm{sp})}_{0}(v),\; \mathbf{T}^{(\ell,\mathrm{tmp})}_{0}(v) \in \mathbb{R}^{C_{\mathrm{VGGT}}\times H'\times W'}$. We then concatenate along the embedding dimension to form the spatialtemporal teacher feature used for distillation:
\begin{equation}
\mathbf{T}^{(\ell)}_{0}(v) = \big[\mathbf{T}^{(\ell,\mathrm{sp})}_{0}(v)\ ;\ \mathbf{T}^{(\ell,\mathrm{tmp})}_{0}(v)\big]
\! \in\! \mathbb{R}^{(2C_{\mathrm{VGGT}})\times H'\times W'},
\end{equation}
where $H',W'$ are the patch-grid height and width. We then use a teacher projector to convert the teacher's embedding into a compact aligned feature dimension $C_a$, and bilinearly upsample the patch-grid to the rendered map size,
$$
    \mathbf{T'}^{(\ell)}_{0}(v) = \mathrm{Projector_{T}}(\mathbf{T}^{(\ell)}_{0}(v))\! \in\! \mathbb{R}^{C_a\times H\times W}.
$$

\noindent\textbf{4D Feature Distillation.}
Given the Gaussian set $\mathcal{G}_0$ of the reference frame, we first render the corresponding student features $\mathbf{S}_{0}$ by projecting the original Gaussian embedding dimension $C_g$ to the aligned compact dimension $C_a$,
$$
    \mathbf{S}_{0} = \mathrm{Projector_{S}}(\mathcal{G}_0)\! \in\! \mathbb{R}^{C_a\times N}.
$$

After that, we render a per-pixel student feature map by GS in each camera view:
\begin{equation}
\mathbf{S'}_{0}(v)=\mathrm{Render}\big(\mathbf{S}_{0}(v);\, \mathbf{K}_v,\mathbf{E}_v\big)\in\mathbb{R}^{C_a\times H\times W},
\end{equation}
where $\mathbf{K}_v$ and $\mathbf{E}_v$ are the intrinsics and extrinsics, and $H$ and $W$ are height and width of rendered student feature map. 
Each Gaussian contributes to the image plane by its $\boldsymbol{\mu}_i(0)$, $\mathbf{r}_i(0)$, $\mathbf{s}_i(0)$, and $\mathbf{\alpha}_i(0)$.

We compute a cosine similarity loss at the reference frame on the selected block $\ell$ over all cameras:
\begin{equation}
\mathcal{L}_\mathrm{distill}
=\frac{1}{|\mathcal{V}||\Omega|}\sum_{v\in\mathcal{V}}\!
\sum_{u\in\Omega}\Big(1-\cos\!\big(\mathbf{T'}^{(\ell)}_{0}(v)[u],\,\mathbf{S'}_{0}(v)[u]\big)\Big),
\end{equation}
where $u$ indexes spatial locations on the feature grid $\Omega$.

\subsection{Optimization and Inference}
\label{subsec:supervision}
Following GaussianFlowOcc, we obtain 2D supervision from pseudo labels generated by Grounded-SAM and Metric3D. 
Given the rendered GS outputs at each view $v$, we compute a pixel-wise segmentation loss $\mathcal{L}_{\text{seg}}$ and a depth regression loss $\mathcal{L}_{\text{dep}}$:
\begin{align}
\mathcal{L}_{\text{seg}} &= 
\frac{1}{|\mathcal{V}||\Omega|}
\sum_{v\in\mathcal{V}}\sum_{u\in\Omega}
\mathrm{CE}\!\left(
\hat{y}_{v}(u),\, y^{\text{pseudo}}_{v}(u)
\right), \label{eq:l_seg}\\
\mathcal{L}_{\text{dep}} &= 
\frac{1}{|\mathcal{V}||\Omega|}
\sum_{v\in\mathcal{V}}\sum_{u\in\Omega}
\left|
\hat{d}_{v}(u) - d^{\text{pseudo}}_{v}(u)
\right|, \label{eq:l_depth}
\end{align}
where $\Omega$ indexes image pixels, 
$\hat{y}_v(u)$ and $\hat{d}_v(u)$ are the rendered semantic and depth predictions, 
and $y^{\text{pseudo}}_v(u)$, $d^{\text{pseudo}}_v(u)$ are pseudo ground-truths.

Together with the distillation and deformation objectives. The final loss is the weighted sum of all components:

\begin{equation}
\mathcal{L}_{\text{total}} =
\lambda_{\text{seg}}\mathcal{L}_{\text{seg}} +
\lambda_{\text{dep}}\mathcal{L}_{\text{dep}} +
\lambda_{\text{distill}}\mathcal{L}_{\text{distill}} +
\lambda_{\text{def}}\mathcal{L}_{\text{def}},
\label{eq:l_total}
\end{equation}
where $\lambda_{\text{seg}}$, $\lambda_{\text{dep}}$, 
$\lambda_{\text{distill}}$, and $\lambda_{\text{def}}$ balance the four terms. 
This enables the network to jointly learn 
geometry, semantics, and motion in a unified weakly supervised setting.

During inference time, only a single frame of multi-view images is provided.
The network predicts a single set of 3D Gaussians $\{G_i(0)\}$ for the current time step, 
which are then voxelized into an occupancy grid for downstream 3D scene understanding.
Thus, temporal deformation is used to \emph{learn} a motion-aware prior, 
while the final deployment operates in a per-frame feed-forward manner without temporal input.
\section{Experiments}
\label{sec:exp}

\definecolor{barrier}{RGB}{255, 106, 30}
\definecolor{bicycle}{RGB}{255, 188, 203}
\definecolor{bus}{RGB}{255, 255, 0}
\definecolor{car}{RGB}{255, 0, 0}
\definecolor{construct}{RGB}{0, 255, 255}
\definecolor{motor}{RGB}{205, 179, 0}
\definecolor{pedestrian}{RGB}{0, 155, 249}
\definecolor{traffic}{RGB}{255, 239, 144}
\definecolor{trailer}{RGB}{148, 52, 0}
\definecolor{truck}{RGB}{177, 5, 244}
\definecolor{driveable}{RGB}{255, 0, 255}
\definecolor{other}{RGB}{139, 137, 137}
\definecolor{sidewalk}{RGB}{85, 0, 76}
\definecolor{terrain}{RGB}{110, 243, 65}
\definecolor{manmade}{RGB}{230, 230, 251}
\definecolor{vegetation}{RGB}{0, 178, 0}
\definecolor{others}{RGB}{0, 0, 0}

\begin{table*}[t]
    \centering
    \newcommand{\clsname}[2]{
        \multicolumn{1}{c}{
            \rotatebox{90}{
                \hspace{-6pt}
                \textcolor{#2}{$\blacksquare$} #1
            }}}
    \setlength{\tabcolsep}{2.08pt}
    \renewcommand\arraystretch{1.05}

    \resizebox{\textwidth}{!}{
    \begin{tabular}{l|cccc>{\columncolor[gray]{0.85}}c|ccccccccccccccc}
        \toprule
        \textbf{Method}
        & \multicolumn{1}{c}{\textbf{IoU}}
        & \textbf{InsM}
        & \textbf{ScnM}
        & \textbf{HCM}
        & \textbf{mIoU}

        & \clsname{bicycle}{bicycle}
        & \clsname{motorcycle}{motor}
        & \clsname{pedestrian}{pedestrian}
        & \clsname{bus}{bus}
        & \clsname{car}{car}
        & \clsname{cons. veh.}{construct}
        & \clsname{trailer}{trailer}
        & \clsname{truck}{truck}

        & \clsname{barrier}{barrier}
        & \clsname{traffic cone}{traffic}
        & \clsname{drive. surf.}{driveable}
        & \clsname{sidewalk}{sidewalk}
        & \clsname{terrain}{terrain}
        & \clsname{manmade}{manmade}
        & \clsname{vegetation}{vegetation} \\
        \midrule
        
        DistillNeRF~\cite{wang2024distillnerf}
        & 29.11 & 4.78 & 20.80 & 3.20 & 10.12
        & 2.08 & 1.98 & 5.54 & 10.21 & 10.09 & 2.56 & 1.43 & 7.90
        & 1.35 & 4.62 & 43.02 & 16.86 & 15.02 & 14.06 & 15.06 \\

        SelfOcc~\cite{huang2024selfocc}
        & 45.01 & 3.00 & 25.63 & 1.19 & 10.54
        & 0.66 & 0.80 & 2.10 & 5.46 & 12.54 & 0.00 & 0.00 & 8.25
        & 0.15 & 0.00 & 55.49 & 26.30 & 26.54 & 14.22 & 5.60 \\

        OccNeRF~\cite{zhang2025occnerf}
        & 22.81 & 3.24 & 25.96 & 1.38 & 10.81
        & 0.82 & 0.23 & 3.10 & 5.13 & 12.49 & 3.50 & 0.52 & 3.90
        & 0.83 & 1.84 & 52.62 & 20.81 & 24.75 & 18.45 & 13.19 \\

        GaussianOcc~\cite{gan2025gaussianocc}
        & - & 6.77 & 22.85 & 5.53 & 11.26
        & 5.82 & 2.82 & 7.95 & 14.58 & 13.55 & 1.30 & 0.56 & 9.61
        & 1.79 & 9.76 & 44.59 & 20.10 & 17.58 & 8.61 & 10.29 \\

        LangOcc~\cite{boeder2024langocc}
        & - & 7.02 & 22.08 & 8.13 & 12.04
        & 7.20 & 10.80 & 6.40 & 5.80 & 13.90 & 0.50 & 3.20 & 11.00
        & 2.70 & 8.70 & 42.10 & 12.50 & 27.20 & 14.10 & 14.50 \\

        GaussTR~\cite{jiang2025gausstr}
        & \underline{45.19} & 7.79 & 24.21 & 5.81 & 13.26
        & 5.22 & 7.08 & 5.12 & 14.07 & \textbf{20.43} & \underline{5.70} & 0.92 & \textbf{13.36}
        & 2.09 & 3.93 & 39.44 & 15.68 & 22.89 & \underline{21.17} & \underline{21.87} \\

        VEON~\cite{zheng2024veon}
        & - & 7.15 & 27.64 & 3.73 & 13.95
        & 2.70 & 3.80 & 4.70 & 14.70 & 10.90 & \textbf{11.00} & 5.30 & 9.60
        & 4.80 & 4.00 & 46.50 & 21.10 & 22.10 & \textbf{24.80} & \textbf{23.70} \\

        GaussianFlow*~\cite{boeder2025gaussianflowocc}
        & 40.39 & \underline{9.59} & \underline{29.62} & \underline{9.73} & \underline{16.27}
        & \underline{9.81} & 10.70 & \underline{8.68} & \underline{17.41} & 15.93 & 4.01 & 0.79 & 11.87
        & \textbf{6.89} & \underline{9.84} & \underline{59.17} & \underline{29.94} & \underline{32.20} & 14.41 & 12.36 \\

        DeGO (ours)
        & \textbf{45.38} & \textbf{10.34} & \textbf{33.46} & \textbf{11.04} & \textbf{18.05}
        & \textbf{10.68} & \textbf{12.88} & \textbf{9.56} & \textbf{18.15} & \underline{17.55} & 3.83 & 0.95 & \underline{12.36}
        & \underline{6.63} & \textbf{10.80} & \textbf{66.19} & \textbf{34.71} & \textbf{37.69} & 15.71 & 12.98 \\
        
        \bottomrule
    \end{tabular}}
    
    \caption{\textbf{Quantitative comparison of weakly supervised methods on the Occ3D-NuScenes~\cite{caesar2020nuscenes,tian2023occ3d} benchmark.} We report the overall IoU and mIoU, as well as Instance mIoU (InsM), Scene mIoU (ScnM), and the proposed Human-centric mIoU (HCM) that focuses on safety-critical human classes. * indicates reproduced results from official code. `cons. veh.' is short for construction vehicle, and `drive. surf.' indicates driving surface. \textbf{Best} number for each class is in bold, and the \underline{second best} is underlined.}
    \label{table:main_result}
\end{table*}

\noindent\textbf{Dataset.} We conduct experiments on the widely used Occ3D-NuScenes benchmark~\cite{tian2023occ3d, caesar2020nuscenes}, which contains 40k frames, each with six camera views. The 3D scene is represented as a voxel grid of size [200, 200, 16] along the x–y–z axes, with a voxel size of 0.4 meters. Following~\cite{huang2024selfocc, jiang2025gausstr, gan2025gaussianocc}, we use 700 scenes for training and 150 for validation.

\noindent\textbf{Metrics.} For evaluation, we use standard metrics~\cite{huang2024selfocc} including IoU and mIoU. We also report the Instance mIoU (InsM) and Scene mIoU (ScnM)~\cite{liu2025disentangling} to check how well the model performs on both object-level and scene-level understanding. In addition, we propose a Human-centric mIoU (HCM) to focus on safety-critical nonrigid classes, including pedestrians, bicycles, and motorcycles.

\subsection{Comparison with State-of-the-Art}
\Cref{table:main_result} compares recent weakly supervised methods on the Occ3D-NuScenes benchmark. DeGO achieves the highest overall performance, surpassing all baselines in both mIoU and IoU. Specifically, compared to the previous model with the best mIoU, DeGO improves mIoU by \textbf{10.9\%} and IoU by \textbf{12.4\%}, demonstrating stronger voxel-level semantic and geometric understanding. GaussianFlow~\cite{boeder2025gaussianflowocc} achieves a balanced overall accuracy across categories, but falls short in handling fine-grained deformation.
In contrast, DeGO benefits from the proposed decoupled Gaussian deformation and foundation knowledge distillation, achieving strong performance on both instance- and scene-level metrics. Notably, it improves HCM by 13.5\% and InsM by 7.8\%, demonstrating enhanced understanding of both nonrigid and rigid objects. Furthermore, the 13.0\% increase in ScnM underscores the model’s ability to represent both dynamic agents and static scene structures robustly.

\subsection{Ablation Study}
\begin{table}[t]
\centering
\resizebox{.5\textwidth}{!}{ 
\begin{tabular}{lcccc}
\toprule
 \textbf{Deformation}& \multicolumn{2}{c}{\textbf{Foundation Model Distillation}} &  \\
\textbf{Module} & \textbf{DINOv2~\cite{oquab2023dinov2}} & \textbf{VGGT}~\cite{wang2025vggt} &  \textbf{mIoU} & \textbf{IoU}\\
\midrule
\xmark  & \xmark & \xmark & 12.06  & 36.41\\
\xmark & \xmark & \cmark & 12.26 & 36.54 \\
\cmark & \xmark & \xmark & 17.29 & 43.67\\
\cmark & \cmark & \xmark & 17.35 & 43.15\\
\cmark & \xmark & \cmark & \textbf{18.05} & \textbf{45.38}\\
\bottomrule
\end{tabular}
}
\caption{\textbf{Ablation of the deformation module and different foundation-model distillation settings.}}
\label{tab:framework}
\end{table}
We analyze the contribution of each component in our framework. 
As shown in \Cref{tab:framework}, introducing the deformation module leads to a substantial 43.4\% improvement over the baseline, 
highlighting its effectiveness in modeling dynamic geometry. 
Building on this, incorporating DINOv2~\cite{oquab2023dinov2} distillation provides a modest gain by aligning Gaussian features with per-image representations. 
In contrast, employing VGGT~\cite{wang2025vggt} as the teacher yields the highest performance, achieving an additional 4.4\% improvement. 
We attribute this to VGGT’s pre-training on multi-view and temporal data, 
which transfers both cross-camera and cross-frame reasoning abilities to our model, offering more consistent and informative results for 4D Gaussian features. In particular, when the deformation module is disabled, VGGT distillation brings about only moderate gains, indicating that deformation learning and foundation-model distillation are mutually strengthening.

\noindent\textbf{Number of Deformed Frames.}
Since the deformation module operates across the temporal domain, we study how the number of deformed frames affects performance. We vary the temporal horizon and evaluate how long-term deformation influences scene prediction.
\Cref{tab:horizon} shows that increasing the number of deformed frames from 4 to 8 consistently improves performance, reaching the best mIoU of 18.05. However, using longer horizons begins to degrade accuracy, as predicting 12 deformed frames requires modeling future scenes up to 6 seconds ahead, which is a challenging setting for reliable 4D Gaussian prediction. This result suggests that moderate temporal deformation provides sufficient motion context while maintaining stability.
\begin{table}[t]
\centering
\resizebox{.5\textwidth}{!}{ 
\begin{tabular}{lccccc}
\toprule
\textbf{Deformed Frames} & \textbf{4} & \textbf{6} & \textbf{8} & \textbf{10} & \textbf{12} \\
\midrule
\textbf{mIoU} & 16.81 & 17.33 & \textbf{18.05} & 17.69 & 12.48 \\
\textbf{IoU} & 42.69 & 44.88 & \textbf{45.38} & 45.29 & 38.12 \\
\bottomrule
\end{tabular}
}
\caption{\textbf{Impact of the number of deformed frames.}}
\label{tab:horizon}
\end{table}

\noindent\textbf{The Impact of Deformation Parameters.}
\begin{table}[t]
\centering
\resizebox{.5\textwidth}{!}{ 
\begin{tabular}{lccccc}
\toprule
\textbf{Rotation} & \textbf{Scale} & \textbf{Opacity} & \textbf{Rigid Mask} & \textbf{mIoU} & \textbf{IoU}\\
\midrule
\xmark & \xmark & \xmark & \xmark & 12.26 & 36.54 \\
\cmark & \xmark & \xmark & \xmark & 12.77 & 38.39 \\
\cmark & \cmark & \xmark & \xmark & 17.06 & 44.50 \\
\cmark & \cmark & \cmark & \xmark & 17.43 & 43.78 \\
\cmark & \cmark & \cmark & \cmark & \textbf{18.05} & \textbf{45.38} \\

\bottomrule
\end{tabular}
}
\caption{\textbf{Ablation on different parameters in deformation module.}}
\label{tab:deformation}
\end{table}
To further analyze the contribution of each component in the deformation module, we conduct an ablation on its internal parameters, including rotation, scale, opacity, and the rigid-body mask. 
As shown in \Cref{tab:deformation}, the scale parameter plays the most critical role, as it directly alters the underlying geometry and thus has a strong impact on occupancy quality. 
Introducing opacity refinement further improves performance by providing finer control over spatial blending. 
Finally, incorporating the rigid-body mask yields the highest mIoU and IoU, 
demonstrating the effectiveness of adaptively distinguishing rigid and nonrigid motion.

\noindent\textbf{Distillation with Spatial and Temporal Attention.} To examine the role of spatial and temporal cues in VGGT distillation, we separately evaluate the effects of cross-camera and cross-frame attention. As shown in \Cref{tab:distillation_module}, distilling from cross-camera features improves performance by 2.0\%, while cross-frame features bring a 1.4\% gain. Combining both yields the best result with a total 4.4\% improvement, confirming that integrating spatial and temporal attention provides complementary information and leads to a more consistent scene understanding.

\begin{table}[t]
\centering
\begin{tabular}{lccc}
\toprule
\textbf{Cross-camera Att.} & \textbf{Cross-frame Att.} &  \textbf{mIoU} & \textbf{IoU}\\
\midrule
\xmark  & \xmark & 17.29 & 43.67\\
\cmark & \xmark & 17.64 & 44.38 \\
\xmark & \cmark & 17.54 & 44.84 \\
\cmark & \cmark & \textbf{18.05} & \textbf{45.38} \\
\bottomrule
\end{tabular}
\caption{\textbf{Ablation on cross-camera and cross-frame distillation.}}
\label{tab:distillation_module}
\end{table}

\noindent\textbf{Effect of Different Teacher Layers.} The VGGT~\cite{wang2025vggt} teacher network contains 24 Transformer blocks in total. To understand how different layers contribute to our task, we select several representative layers from the early, middle, and late stages, and distill features from each individually.
From \Cref{tab:distillation_layer}, we find that deeper layers generally provide richer semantic information, leading to consistent improvements in mIoU from shallow to deep layers. However, we observe a performance drop when it comes to the last layer (index=23) as the last layer of the foundation model is more task-specific, and it will hinder the distillation gain since a lack of geometric knowledge.

\begin{table}[t]
\centering
\resizebox{.5\textwidth}{!}{ 
\begin{tabular}{lccccc}
\toprule
{} & \textbf{Shallow Layer} & \textbf{Middle Layer} &  \multicolumn{3}{c}{\textbf{Deep Layer}}\\
\midrule
\textbf{Layer Index} & 8 & 13 & 20 & 22 & 23\\
\textbf{mIoU} & 17.48 & 17.8 & 17.95 & \textbf{18.05} & 17.88\\
\textbf{IoU} & 44.62 & 44.93 & 45.18 & \textbf{45.38} & 43.71\\
\bottomrule
\end{tabular}
}
\caption{\textbf{Impact of different teacher layers on distillation.}}
\label{tab:distillation_layer}
\end{table}

\noindent\textbf{Where to Apply the Distillation.} As illustrated in \Cref{fig:model}, our distillation is applied to the Gaussian Transformer, which intends to transfer knowledge from 4D foundation knowledge to the Gaussian features. An alternative choice is the image encoder, which also lacks cross-camera and cross-frame awareness. However, as shown in our experiments, distilling from the image encoder results in an mIoU of 17.11, dropping about 5.2\% mIoU compared to distilling from the Gaussian Transformer. This shows that 3D Gaussian features are more compatible with the VGGT teacher, enabling the student to absorb better spatialtemporal knowledge than with 2D image features.

\noindent\textbf{Teacher-Student Projection Dimension.} We also investigate the role of the projection dimension in distillation, which controls the expressiveness of the teacher–student alignment space. \Cref{tab:projection_dimension} shows that increasing the projection dimension from 16 to 32 improves mIoU from 17.15 to 18.05, while a larger dimension of 48 slightly reduces performance. This suggests that a moderate projection dimension provides sufficient capacity for effective teacher–student alignment while avoiding redundancy and unstable optimization. A compact projection space implicitly encourages smoother alignment, leading to more reliable feature transfer across diverse scenes.

\begin{table}[t]
\centering
\begin{tabular}{lccc}
\toprule

\textbf{Projection Dimension} & \textbf{16} & \textbf{32} & \textbf{48} \\
\midrule
\textbf{mIoU} & 17.15 & \textbf{18.05} & 17.35 \\
\textbf{IoU} & 42.99 & \textbf{45.38} & 43.21 \\
\bottomrule
\end{tabular}

\caption{\textbf{Ablation on the projection dimension in the teacher-student alignment module.}}
\label{tab:projection_dimension}
\end{table}

\begin{figure*}[t]
    \centering
    \includegraphics[width=\linewidth]{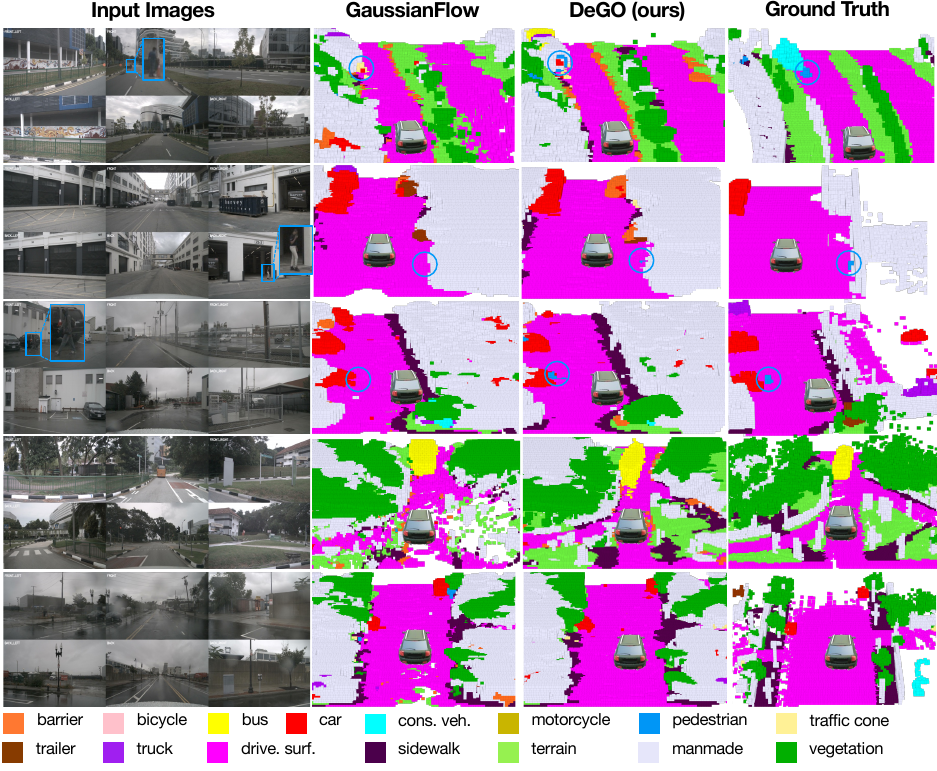}
    \caption{\textbf{Qualitative comparison with the state-of-the-art method.} The upper three scenes focus on Human-centric nonrigid classes, and the lower two scenes focus on static context.}
    \label{fig:vis_all}
\end{figure*}

\subsection{Visualizations}
To qualitatively compare our method with the state-of-the-art GaussianFlow, we visualize voxel predictions and ground truth for several scenes. As shown in \Cref{fig:vis_all}, the top three rows highlight human-centric scenarios, while the bottom two focus on static scene context. Our method shows clear advantages in recognizing the nonrigid pedestrians. For example, in the first scene, the pedestrian is far from the front camera yet still correctly detected by our model. In the third scene, the pedestrian’s clothing color closely matches the background vehicle, but our approach still identifies it accurately. It shows our advantage in human-centric nonrigid motion. For static elements, the last two scenes show that our model better reconstructs driving surfaces and scene boundaries compared with GaussianFlow. Overall, these visual results are consistent with the quantitative findings in \Cref{table:main_result}, confirming stronger performance on both human-centric and static scene elements.

\section{Conclusion}
\label{sec:conclusion}
In this work, we introduced a deformable Gaussian occupancy framework that integrates \textit{decoupled Gaussian deformation} with \textit{foundation knowledge distillation} for dynamic scene understanding. By decoupling nonrigid and rigid motion, our method models human-centric classes while preserving geometric stability for rigid classes. The proposed factorized distillation transfers multi-view and temporal reasoning from foundation models, producing foundation-aligned 4D Gaussian features. Results on Occ3D-NuScenes confirm clear gains over prior methods, with stronger modeling of human motion and stable scene geometry. Future work will explore longer sequences, additional modalities, and large-scale pretraining for broader 4D generalization.

\section{Acknowledgement}
\label{sec:acknowledgement}

The authors would like to thank Lan Feng and Zimin Xia for their valuable feedback. This work was supported by Sportradar\footnote{\href{https://sportradar.com/}{https://sportradar.com/}} (Yang's Ph.D.), Swiss AI Initiative, and Honda R\&D Co., Ltd.

{
    \small
    \bibliographystyle{ieeenat_fullname}
    \bibliography{main}
}

\clearpage
\setcounter{page}{1}
\maketitlesupplementary

\section{Effect of Deformation on Gaussian Sparsity}
\Cref{tab:gaussian_ablation} shows the robustness of our model under reduced Gaussian density by progressively decreasing the number of Gaussian primitives. Through it, we can find the performance degrades more when removing the deformation module, indicating a strong reliance on dense Gaussian representations. In contrast, adding the deformation module yields smaller performance drops across all settings, demonstrating its benefit in improving robustness to sparse Gaussians. This suggests that our method can maintain accurate scene modeling even with fewer primitives, enabling more efficient inference.

\begin{table}[htbp]
\centering
\resizebox{.5\textwidth}{!}{ 

\begin{tabular}{lcccc}
\toprule
 & \textbf{10000} & \textbf{5000} & \textbf{1000} & \textbf{500} \\
\midrule
w/o Deformation Module 
& 12.26 
& 10.17 (-17\%) 
& 8.85 (-28\%) 
& 6.02 (-51\%) \\

w/ Deformation Module 
& \textbf{18.05} 
& \textbf{17.10} (-5\%) 
& \textbf{14.90} (-17\%) 
& \textbf{13.09} (-27\%) \\
\bottomrule
\end{tabular}
}
\caption{\textbf{Performance under Reduced Gaussian Density.}}
\label{tab:gaussian_ablation}
\end{table}

\section{Efficiency}

\begin{figure}[htbp]
    \centering
    \includegraphics[width=\linewidth]{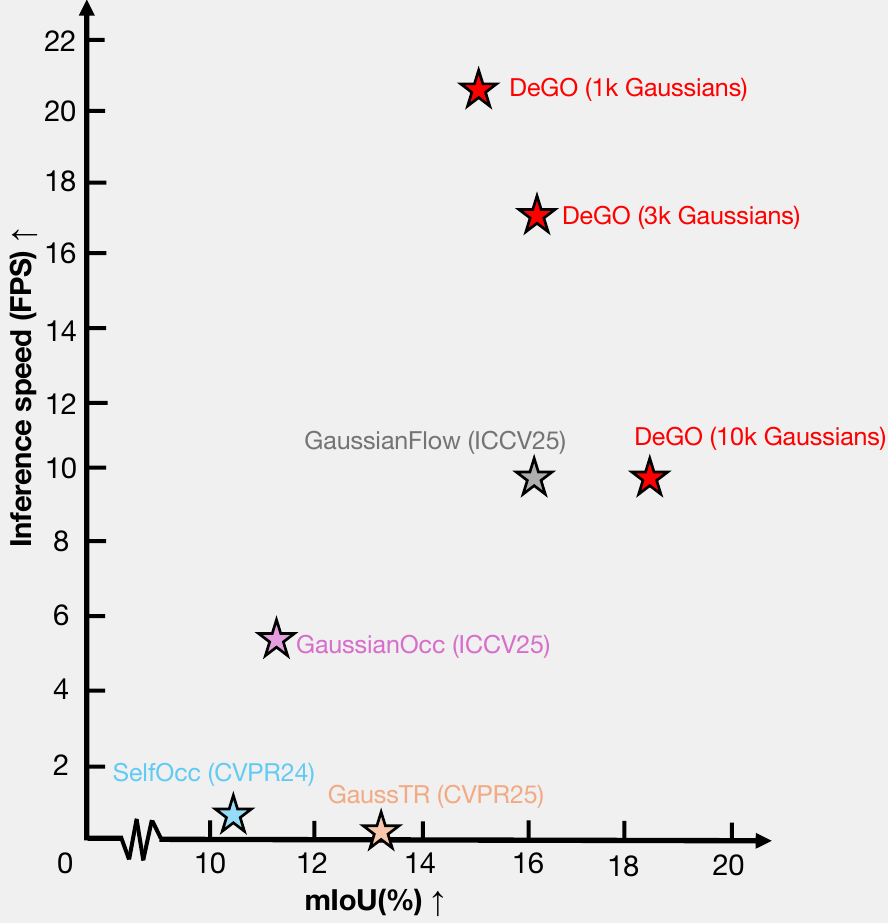}
    \caption{\textbf{Comparison of inference speed and accuracy with state-of-the-art methods on Occ3D-NuScenes validation set.}}
    \label{fig:inference}
\end{figure}

To assess model efficiency, we follow the evaluation protocol of GaussianFlow \cite{boeder2025gaussianflowocc} and measure inference speed on the full Occ3D-NuScenes validation set \cite{caesar2020nuscenes,tian2023occ3d} using a single A100 GPU. As shown in Figure 1, our model achieves over 20 FPS with only 1k Gaussians while maintaining competitive accuracy. With 3k Gaussians, it matches the mIoU of the best prior method while providing over 70\% faster inference. These results demonstrate that our method enables efficient inference with significantly fewer Gaussians.

\section{Performance on Ray-based Metric}
In addition to conventional metrics such as IoU and mIoU for occupancy prediction, we also report results using the ray-based metric RayIoU introduced in \cite{liu2024fully}. Unlike standard voxel-level IoU, the ray-based metric computes the agreement between predicted and ground-truth voxels only along each camera ray. This formulation focuses on view-consistent occupancy and avoids penalizing voxels that are never observed by cameras.

As shown in \Cref{tab:rayIoU}, we follow the evaluation protocol of \cite{boeder2025gaussianflowocc} and compare RayIoU at multiple depth intervals. Our method consistently surpasses previous state-of-the-art baselines across all depth units, demonstrating the strong view consistency of our predicted occupancy.

\begin{table}[t]
\centering
\resizebox{.5\textwidth}{!}{ 
\begin{tabular}{lcccc}
\toprule
\textbf{Method} & \textbf{RayIoU} & \textbf{RayIoU@1} & \textbf{RayIoU@2} & \textbf{RayIoU@4}\\
\midrule
GaussianOcc~\cite{gan2025gaussianocc} & 13.43 & 9.85 & 13.49 & 16.94 \\
GaussianFlow~\cite{boeder2025gaussianflowocc} & 18.00 & 12.24 & 18.13 & 23.69 \\
DeGO (ours) & \textbf{18.89} & \textbf{13.37} & \textbf{18.93} & \textbf{24.37} \\
\bottomrule
\end{tabular}
}
\caption{\textbf{Performance under ray-based metrics.} Similar to the main text, we exclude the `others' and `other flat' classes and compute the RayIoU on the other 15 classes.}
\label{tab:rayIoU}
\end{table}

\section{More Visualizations}
\begin{figure*}[t]
    \centering
    \includegraphics[width=\linewidth]{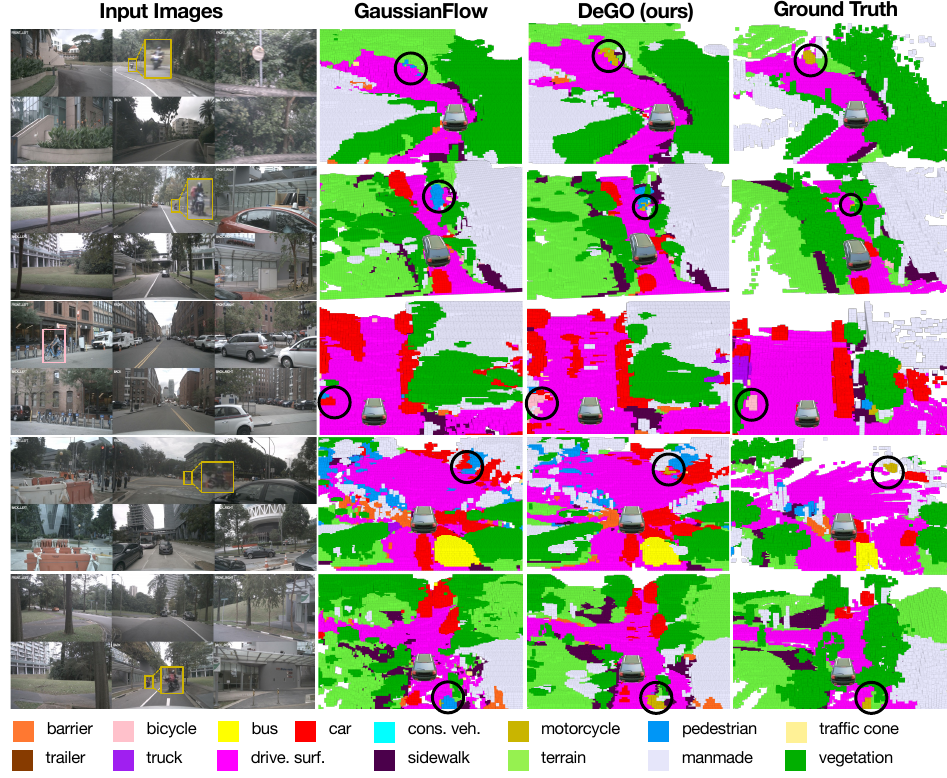}
    \caption{\textbf{Additional qualitative comparison with the state of the art.} We highlight two additional human-centric classes, bicycle and motorcycle. Compared with GaussianFlow, our method produces more accurate predictions for these visually similar categories.}
    \label{fig:app_vis}
\end{figure*}

To further examine performance on deformable, human-centric classes, we provide additional qualitative comparisons of the baseline, our method, and the ground-truth occupancy predictions. As shown in \Cref{fig:app_vis}, the baseline often misclassifies bicycles and motorcycles as pedestrians. Although these categories can appear visually similar, our method more reliably distinguishes them, producing noticeably more accurate predictions.

\section{Implementation Details}

\begin{table*}[t]
\centering
\begin{tabular}{lll}
\toprule
\textbf{Parameter} & \textbf{Value} & \textbf{Description} \\
\midrule
\multicolumn{3}{l}{\emph{Data \& input: }} \\
Number of cameras      & 6                         & Surround-view camera inputs \\
Input image size       & $256 \times 704$         & Network input resolution \\
Source image size      & $900 \times 1600$        & Original image resolution \\
Voxel grid range $(x,y,z)$ & $[-40,40]\times[-40,40]\times[-1,5.4]$ m & 3D occupancy volume \\
Voxel size $(x,y,z)$   & $0.4, 0.4, 0.4$ m        & Spatial resolution of BEV voxels \\

\midrule
\multicolumn{3}{l}{\emph{Model:}} \\
Backbone               & ResNet-50                & Image feature extractor \\
Neck                   & FPN                      & Multi-scale feature aggregation \\
Hidden dimension       & 256                      & BEV and Gaussian feature channels \\
Gaussian init scale    & 4                        & Initial Gaussian scale factor \\
Temporal frame IDs     & $[-8,\dots,8]$          & Context frames for deformation \\
Training clip length   & 8                        & Number of adjacent frames per clip \\
Deformation depth      & 6                        & Layers in deformation MLP \\
Time embedding dim     & 32                       & Temporal embedding dimension \\
Positional enc. levels & 6                        & Levels for 3D position encoding \\
Time enc. levels       & 4                        & Levels for time encoding \\
VGGT teacher layer     & 22                       & VGGT layer index for distillation \\
Gaussian feature dim   & 32                       & Per-Gaussian distilled feature size \\
Teacher feature dim    & 2048                     & VGGT feature dimension \\
Distillation loss      & cosine                   & Normalized cosine feature loss \\
\midrule
\multicolumn{3}{l}{\emph{Training:}} \\
Optimizer              & AdamW                    & Optimization algorithm \\
Learning rate          & $1\times10^{-4}$        & Initial learning rate \\
Weight decay           & $1\times10^{-2}$        & Weight decay factor \\
LR schedule            & cosine w/ warm-up        & Custom cosine annealing schedule \\
Warm-up iterations     & 200                      & Linear warm-up steps \\
Warm-up ratio          & 0.001                    & Warm-up start LR ratio \\
Max epochs             & 30                       & Total training epochs \\
Batch size             & 4                        & Samples per GPU \\
Workers per GPU        & 12                       & Data loading workers \\
Gradient clipping      & 5.0                      & Max gradient norm \\

\bottomrule
\end{tabular}
\caption{\textbf{Detailed hyperparameters used in our implementation.}}
\label{tab:hyperparams}
\end{table*}
We use a ResNet-50~\cite{he2016deep} image encoder and a Gaussian Transformer consisting of three blocks with a hidden dimension of 256. The deformation module is configured with 32 temporal channels, a positional encoding level of 6, and a time-encoding level of 4. The feature network is a 6-layer MLP, and the 4D Gaussian prediction heads are implemented as 2-layer MLPs.
For VGGT distillation, we use the feature layer at index 22 and project it to a 32-dimensional space.
The model is trained with a batch size of four on four A100 GPUs using the Adam optimizer~\cite{loshchilov2017decoupled}, starting with a learning rate of 1e-4 and a decay rate of 1e-2. To ensure stable optimization, we adopt a cosine learning-rate schedule with warm-up and standard gradient clipping. Additional architectural, data processing, and training hyperparameters are summarized in \Cref{tab:hyperparams} for completeness.



\end{document}